\documentclass[twoside]{article}

\usepackage{aistats2020}
\usepackage{bm}
\usepackage{amssymb}
\usepackage{booktabs}
\usepackage{xr}
\usepackage{algorithmic,algorithm}
\usepackage{multirow}
\usepackage{graphicx}

\usepackage{cancel}
\usepackage[dvipsnames]{xcolor}
\usepackage[round]{natbib}

\begin{document}

%

%

\twocolumn[

\aistatstitle{Streaming Adaptive Nonparametric Variational Autoencoder }

\aistatsauthor{ Tingting Zhao \And Zifeng Wang \And  Aria Masoomi  \And Jennifer Dy}

\aistatsaddress{ Northeastern University \And  Northeastern University \And Northeastern University \And Northeastern University} 
]

\begin{abstract}
We develop a data driven approach to perform clustering and end-to-end feature learning simultaneously for streaming data that can adaptively detect novel clusters in emerging data. Our approach, Adaptive Nonparametric Variational Autoencoder (AdapVAE), learns the cluster membership through a Bayesian Nonparametric (BNP) modeling framework with Deep Neural Networks (DNNs) for feature learning. We develop a joint online variational inference algorithm to learn feature representations and clustering assignments simultaneously via iteratively optimizing the Evidence Lower Bound (ELBO). We
resolve the catastrophic forgetting \citep{kirkpatrick2017overcoming} challenges with streaming data by adopting generative samples from the trained AdapVAE using previous data, which avoids the need of storing and reusing past data. We demonstrate the advantages of our model including adaptive novel cluster detection without discarding useful information learned from past data, high quality sample generation and comparable clustering performance as end-to-end batch mode clustering methods on both image and text corpora benchmark datasets.
\end{abstract}

\section{Introduction}
Clustering is an important unsupervised learning problem in machine learning. It is the task of grouping similar objects together such that there is high intra-cluster similarity and low inter-cluster similarity among different objects. Cluster analysis depends on the definition of similarity among objects. Similarity, in turn, depends on the feature space, in which the representation of the data is defined. Clustering is usually carried out in a batch mode with the entire dataset available. However, data streams often arrive in real time and it is desirable to perform clustering without revisiting past data. The model should also be flexible to expand with data size and complexity. 

In this paper, we revisit clustering in a streaming data setting and try to answer the following questions. Without storing and reusing past data, (1) can we develop a data driven approach to perform end-to-end clustering and feature learning simultaneously? (2) can our approach be `incremental' or `online' for streaming data? (3) can we resolve the phenomenon of forgetting what has been learned as new data arrives for Deep Neural Networks (DNNs)? (4) can our model adapt to changes in data distribution if the distribution evolves in time without discarding useful information learned from historical data and detect novel clusters automatically? 

To answer these questions, we take inspirations from DNNs \citep{jiang2017variational}, Bayesian Nonparametric (BNP) models \citep{broderick2013streaming,hughes2013memoized,tank2015streaming, campbell2015streaming} and Lifelong Learning (LL) \citep{chen2016lifelong, shin2017continual} to develop our novel online clustering approach. Our paper provides an end-to-end feature learning and clustering method from raw data. We learn the feature space through DNNs without having to pre-define similarity or feature engineering. It is desirable to have an unsupervised learning algorithm to continually learn and discover novel clusters as it encounters streaming data. BNP models are natural choices as they allow the number of clusters to grow as new data arrives.

We introduce a novel adaptive clustering algorithm AdapVAE for streaming data based on BNP and Variational Autoencoder (VAE), which is a deep learning technique for learning latent representations. AdapVAE enables adaptive modelling of complex real-world distributions via (1)
learning the structure of feature space while clustering through deep representations; (2) detecting novel
clusters adaptively when new data arrives while maintaining previously learned clusters without storing or
reusing past data; and (3) achieving comparable clustering
accuracy if given a batch data setting with state-of-the-art methods which have knowledge about the true
number of clusters.
\section{Related Work}
Recent research work \citep{dilokthanakul2016deep, jiang2017variational,kilinc2018learning} has focused on combining deep generative models to learn good representations of the original data and probabilistic models to conduct clustering analysis. \citet{johnson2016composing}  have proposed a general modeling and inference framework that combines the complementary strengths of probabilistic graphical models and deep learning methods. Deep Embedded Clustering (DEC) \citep{xie2016unsupervised} has focused on simultaneously learning feature representations and cluster assignments using DNNs. DEC has achieved good performance in clustering but it can not model the generative process of the data. To resolve this issue, Variational Deep Embedding (VaDE) \citep{jiang2017variational} has been proposed to combine VAE \citep{kingma2014auto} and a Gaussian Mixture Model (GMM) to learn representations of the data while performing clustering and generating samples. 

However, the latest existing methods share the following limitations: (1) they all have focused on the case of training in a batch mode; (2) the total number of clusters needs to be fixed in advance; (3) they can not detect potential novel clusters when new data arrives based on a trained model from historical data or they are just able to classify all the emerging novelty as an \emph{outlier} class instead of further categorizing them into different clusters according to the characteristics of the data \citep{williams2002comparative, Kodirov_2015_ICCV, amarbayasgalan2018unsupervised, masana2018metric}; (4) they can not adapt to the data size and complexity automatically. 

BNP models are natural choices as they allow the number of clusters to grow as new data arrives. There have been research work on streaming variational inference for BNP mixture models \citep{tank2015streaming,huynh2016streaming}. \citet{campbell2015streaming} have developed streaming distributed scalable variational inference for BNP to address parallelization. \citet{hughes2013memoized} have developed memoized online variational inference strategies that use birth and merge moves to adapt to data complexity and escape from local maximum. However,
these methods have focused on the original data space, which may be high-dimensional. 

\citet{goyal2017nonparametric} proposed a hierachical nonparametric VAE via combining tree-structured BNP priors with VAEs to enable infinite flexibility of the latent representation space. They developed a variational inference algorithm to learn the DNN and BNP parameters jointly through alternating optimization for a {\em batch setting}. In contrast, our work can learn in a {\em streaming setting}.
Additional challenges occur for streaming data compared with a batch setting. Neural networks tend to forget the information learned in the previously trained tasks using past data when training on new tasks, which leads to degradation performance for past tasks if not storing or reusing previous data. This phenomenon has been referred to as catastrophic forgetting \citep{mccloskey1989catastrophic, kirkpatrick2017overcoming} in the DNN and LL literature. One of the main objectives for LL is to perform incremental learning of new tasks without forgetting previously learned information using past data. 

Our work fills the gap in the literature by developing a Streaming Adaptive Nonparametric Variational Autoencoder (AdapVAE) modelling framework, which extends VAE by: (1) enabling it to perform clustering via Dirichlet Processes Mixture Models (DPMMs) for the latent representation $\bm{z}$; (2) allowing it to detect new clusters to accommodate with increasing emerging data complexity or merge clusters to remove redundancy in a sequential updating process using birth and merge moves; (3) overcoming catastrophic forgetting in a streaming setting without storing or reusing past data via generating a small number of samples based on the current parameter estimates and merge the generated samples with new data to sequentially update the DNN and DPMM parameters; (4) tailoring the algorithm to a streaming setting by exploiting the sequential nature of Bayes theorem to recursively update an approximation of the posterior and update the local and global parameters of DPMMs using different hierarchies of sufficient statistics incrementally.

\section{Problem Statement}

Let $\bm{x}_n$ denote the $n$th observation, for $n=1, 2, \ldots, N$, where $N$ denotes the total number of observations. Given unlabeled data $\bm{x}_n\in X$, where $X$ represents the data space, it is our interest to learn a low-dimensional latent representation for $\bm{x}_n$ while simultaneously clustering the set of $N$ observations in the latent space. For example, $\bm{x}_n$ can represent a hand-written digit image of pixels and we target to group images of the same digit together based on its low-dimensional latent representation $\bm{z}_n$. Unlike existing methods \citep{dilokthanakul2016deep, xie2016unsupervised, jiang2017variational} that fix the number of clusters as the true one, our method is nonparametric, starts with one cluster and can grow or merge clusters to accommodate with data complexities.  

\subsection*{Streaming Data Setting Assumptions}\label{sec:assumption}
Usually, online learning refers to updating the estimates as each single data instance arrives. Our online learning setting has the following assumptions:

\begin{itemize}
\setlength{\parskip}{0pt}
\setlength{\itemsep}{0pt plus 1pt}
\item The data stream arrives in sequential order and each time only one data stream can fit in memory. 
\item When one data stream is in memory, there is no access to its predecessor and successor stream. 
\item Any subset of the current stream in memory can be visited multiple times and we call the subset within each stream as mini-batch. 
\item Once the successor of the current data stream arrives, the stream in memory is no longer available.
\end{itemize}

\subsection{Review on Dirichlet Process} 
The Dirichlet process (DP) is a random probability measure that can be used as a nonparametric prior. It can be seen as a countably infinite sum of atomic measures, where each partition is assigned with an independent parameter from a common distribution. A constructive definition of DP via a stick-breaking process was provided by \citet{sethuraman1982convergence}, which is reviewed below.  

A DP is characterized by a base distribution $G_0$ and a parameter $\alpha$ and is denoted as $\textrm{DP}(G_0, \alpha)$. A stick-breaking prior is of the form $G(\cdot)=\sum_{k=1}^{\infty}\pi_k \delta_{\theta_k}$,
where $\delta_{\theta_k}$ is a discrete measure concentrated at $\theta_k\sim G_0$, which is a random sample from the base distribution $G_0$ \citep{ishwaran2001gibbs} and can be seen as the parameters of the component distribution of a mixture of distributions with mixing proportion $\pi_k$. The $\pi_k$s are random weights independent of $G_0$ but satisfy $0\leqslant \pi_k\leqslant 1$ and $\sum_{k=1}^{\infty} \pi_k=1$. The weights $\pi_k$ can be drawn through an iterative process:
\vspace{-2mm}
\[
\pi_k=\begin{cases}
v_1, & \textrm{if}\quad k=1, \\
v_k\prod_{j=1}^{k-1}(1-v_j), & \textrm{for}\quad k>1,
\end{cases}
\]
where $v_k\sim \textrm{Beta}(1, \alpha)$ so that we obtain the stick-breaking construction of $\textrm{DP}(G_0, \alpha)$. Assume that the latent representation $\bm{z}$ for the raw data $\bm{x}$ comes from a mixture of Gaussian distribution and the number of mixture components is infinite. The DPMM is an appropriate choice for streaming data since if more data is observed, DPMMs allow the number of clusters to grow without bounds. 

\subsection{Review on Variational Autoencoder}\label{sec:vae}
 In order to learn a low-dimensional representation of the data space $X$ while maintaining high quality reconstruction power, we choose VAE \citep{kingma2014auto} as the backbone of our algorithm. VAE assumes the generative model $p_\theta(x|z)$ is parameterized by $g_\theta: Z\rightarrow X$,
 which represents the decoder and maps the latent space $Z$ to the original data space $X$. Similarly, VAE assumes $q_{\psi}(\bm{z}|\bm{x})$ as the variational approximation of the posterior parameterized by $f_\psi: X \rightarrow Z$. Both $g_\theta$ and $f_\psi$ are often chosen as DNNs due to its powerful function approximation \citep{hornik1991approximation} and good feature learning capabilities \citep{kingma2014semi, nalisnick2016approximate}. 
 
 In VAE, it is assumed that $q_{\psi}(\bm{z}|\bm{x})=\mathcal{N}(\bm{\mu}, \mathrm{diag}(\bm{\sigma}^2))$ and $f_\psi(\bm{x})=(\bm{\mu}, \log\bm{\sigma}^2)$, where $f_\psi(\bm{x})$ is a neural network with $(\bm{\mu}, \log\bm{\sigma}^2)$ as output, where both $\bm{\mu}$ and $ \log\bm{\sigma}^2$ depend on $\bm{x}$. The parameters $\theta$ and $\psi$ are learned by minimizing the reconstruction error, which is equivalent to maximizing $\mathbb{E}_{z\sim q_{\psi}(\bm{z}|\bm{x})}\left(\log p_\theta(\bm{x}|\bm{z})\right).$ 
 VAE has good feature learning capacities but it is not able to perform clustering tasks. 
 
 \section{Generative Process of AdapVAE}\label{sec:model}
 We introduce the generative process of AdapVAE. Assume that the latent representation of the observation is a realization from a DP Guassian Mixture,
 \begin{center}
$ G\sim\mbox{DP}(\alpha_0, G_0),\quad G_0=\mbox{NW}(\bm{\lambda}_0)$
    $G\overset{\Delta}{=}\sum_{i=1}^{\infty} w_k\delta_{\phi_k},\quad\phi_{k}=\left(\bm{\mu}_k^*, \bm{\sigma}_k^*\right)\sim\mbox{NW}(\bm{\lambda}_0),
$
\end{center}
\vspace{-1mm}
where we assume that the component covariance $\bm{\sigma}_k^*$ is diagonal and $\mbox{NW}$ represents the Normal-Wishart distribution with parameters $\bm{\lambda}_0$, which is assumed as the base distribution $G_0$. The NW hyper-parameters $\bm{\lambda}_0 = (m_0, \beta_0, \nu_0, W_0)= (\bm{0}, 0.2, D+2, I_{D\times D})$, where $D$ is the dimension of of the latent vector $\bm{z}$, which is a low-dimensional representation of the original data and is generated in step $(b)$ of the generative process. The data can be described as generated from the following process:
\vspace{-1mm}
\begin{itemize}
  \setlength{\parskip}{0pt}
  \setlength{\itemsep}{0pt plus 1pt}
    \item Draw $V_i|\alpha_0\sim\mbox{Beta}(1, \alpha_0)$, $i=\{1, 2, \ldots\}$.
    \item Draw $\left(\bm{\mu}_k^*, \bm{\sigma}_k^*\right)|G_0 \sim G_0$, $k=\{1, 2, \ldots\}$. 
    \item For the $n$th data point $\bm{x}_n$:
        \begin{itemize}
        \item[(a)] Draw a cluster membership $Y_n \sim \mbox{Cat}(\pi(\bm{v}))$, where \[
\pi_k=\begin{cases}
v_1, & \textrm{if}\quad k=1, \\
v_k\prod_{j=1}^{k-1}(1-v_j), & \textrm{for}\quad k>1.
\end{cases}
\]
    \item[(b)] Draw a latent representation vector $Z_n|Y_n=k\sim \mathcal{N}\left(\bm{\mu}_k^*, \bm{\sigma}^{*2}_k\right)$.
    \item[(c)] Generate the $n$th observation $\bm{x}_n$ from $X_n|Z_n=\bm{z}_n \sim \mathcal{N}\left(\bm{\mu}(\bm{z}_n; \theta), \bm{\sigma}^2(\bm{z}_n, \theta)\right)$, where we have
    $\left(\bm{\mu}(\bm{z}_n; \theta), \bm{\sigma}^2(\bm{z}_n, \theta)\right)=g_\theta(\bm{z}_n; \tau)$ as a decoder network to reconstruct the observation $\bm{x}_n$ from the latent vector $\bm{z}_n$. 
    \end{itemize}
\end{itemize}
\vspace{-3mm}
The corresponding joint probability density is 
\begin{align*}
    p(\bm{x}, \bm{y}, \bm{z}, \bm{\phi}, \bm{v})&=p(\bm{x}|\bm{z};\tau)p(\bm{z}\vert \bm{y})p(\bm{y}|\bm{v})p(\bm{v})G_0(\bm{\phi}|\bm{\lambda}_0)\\
    &=\prod_{n=1}^N \mathcal{N}(\bm{x}_n|\bm{\mu}(\bm{z}_n;\theta),\mathrm{diag}(\bm{\sigma}^2(\bm{z}_n, \theta)))\\
    &\quad\,\prod_{k=1}^{\infty}\mathcal{N}(\bm{z}_n|\bm{\mu}_{k}^*, \bm{\sigma}_k^2\bm{I})P(Y_n=k|\pi(\bm{v}))\\
    &\quad\quad\quad \mbox{Beta}(v_k|1, \alpha_0)G_0(\phi_{k}|\lambda_0).
\end{align*}

\section{Streaming Variational Inference for AdapVAE}\label{sec:method}
\subsection{ELBO Derivation}
Since the posterior distribution under DPMMs is intractable, approximate inference methods are required. Variational inference provides an approximation for the posterior $p(\bm{y}, \bm{z}, \bm{\phi}, \bm{v}|\bm{x})$ by casting inference as an optimization problem. It aims to find a surrogate distribution $q(\bm{y}, \bm{z}, \bm{\phi}, \bm{v}|\bm{x})$ that is the most similar to the distribution $p(\bm{y}, \bm{z}, \bm{\phi}, \bm{v}|\bm{x})$ of interest over a class of tractable distributions. 

Given the generative process in Section~\ref{sec:model}, the marginal log-likelihood for data $\bm{x}$ is 
\begin{align*}
    \log p(\bm{x}) &= \log\int_{\bm{z}}\sum_{\bm{y}}\int_{\bm{v}}\int_{\bm{\phi}}p(\bm{x}, \bm{y}, \bm{z},\bm{\phi}, \bm{v}) d\bm{z}d\bold{v}d\bm{\phi},
\end{align*}   
Using Jensen’s inequality, we  obtain 
\begin{align}
\log p(\bm{x}) &\geqslant \mathbb{E}_{q(\bm{y}, \bm{z}, \bm{\phi}, \bm{v}|\bm{x})}\left\{
    \log\frac{p(\bm{x}, \bm{y}, \bm{z}, \bm{\phi}, \bm{v})}{q(\bm{y}, \bm{z}, \bm{\phi}, \bm{v}|\bm{x})}\right\}\\
    &=\mathcal{L}_{\textrm{ELBO}}(\bm{x}),\notag
\end{align}
where $q(\bm{y}, \bm{z}, \bm{\phi}, \bm{v}|\bm{x})$ is the variational posterior distribution used to approximate the true posterior $p(\bm{y}, \bm{z}, \bm{\phi}, \bm{v}|\bm{x})$ and $\mathcal{L}_{\textrm{ELBO}}$ is the Evidence Lower Bound (ELBO). Minimizing the Kullback-Leibler (KL) divergence between $q(\bm{y}, \bm{z}, \bm{\phi}, \bm{v}|\bm{x})$ and $p(\bm{y}, \bm{z}, \bm{\phi}, \bm{v}|\bm{x})$ is equivalent to maximizing the ELBO. We assume $q(\bm{y}, \bm{z}, \bm{\phi}, \bm{v}|\bm{x})$ can be factorized as $q_{\psi}(\bm{z}|\bm{x})q(\bm{y})q(\bm{v})q(\bm{\phi})$. Thus, the ELBO is 

\begin{align} \label{eq:all_elbo}
            \begin{split}
&\mathbb{E}_{q(\bm{y}, \bm{z}, \bm{\phi}, \bm{v}|\bm{x})}
    \left[\log\frac{p(\bm{x}|\bm{z})p(\bm{z}|\bm{y}, \bm{\phi})p(\bm{y}|\bm{v})p(\bm{v})p(\bm{\phi})}{q_{\psi}(\bm{z}|\bm{x})q(\bm{y}|\bm{x})q(\bm{v})q(\bm{\phi}|\bm{x}))}\right]
    \\
    &= \mathbb{E}_{q(\bm{y}, \bm{z}, \bm{\phi}, \bm{v}|\bm{x})}
    \left[\log p(\bm{x}|\bm{z})\right]\\
    &+\mathbb{E}_{q(\bm{y}, \bm{z}, \bm{\phi}, \bm{v}|\bm{x})}
    \left[\log p(\bm{z}|\bm{y}, \bm{\phi})\right]\\
    &- \mathbb{E}_{q(\bm{y}, \bm{z}, \bm{\phi}, \bm{v}|\bm{x})}
    \left[\log q_{\psi}(\bm{z}|\bm{x})\right] 
    \\
   &+ \mathbb{E}_{q(\bm{y}, \bm{z}, \bm{\phi}, \bm{v}|\bm{x})}
    \left[\log p(\bm{y}|\bm{v})\right] 
    + \mathbb{E}_{q(\bm{y}, \bm{z}, \bm{\phi}, \bm{v}|\bm{x})}
    \left[\log p(\bm{v})\right] 
    \\
    &- \mathbb{E}_{q(\bm{y}, \bm{z}, \bm{\phi}, \bm{v}|\bm{x})}
    \left[\log q(\bm{y}|\bm{x})\right] 
    - \mathbb{E}_{q(\bm{y}, \bm{z}, \bm{\phi}, \bm{v}|\bm{x})}
    \left[\log q(\bm{v})\right] 
    \\
    &- \mathbb{E}_{q(\bm{y}, \bm{z}, \bm{\phi}, \bm{v}|\bm{x})}
    \left[\log q(\bm{\phi}|\bm{x})\right] 
    + \mathbb{E}_{q(\bm{y}, \bm{z}, \bm{\phi}, \bm{v}|\bm{x})}
    \left[\log p(\bm{\phi})\right] 
        \end{split}
        \end{align}

Inspired by \citet{goyal2017nonparametric}'s work, we adopt alternating optimization strategy to update the DNN and DPMM parameters by maximizing the ELBO. We update the VAE parameters ($\theta$ and $\psi$) and the latent variable ($\bm{z}$) given the current DPMM parameters. Then, we fix the DNN parameters and update the DPMM parameters (described in Table~\ref{tab:notation}). This strategy allows us to use improved latent representation to infer the clustering structure via DPMMs and the updated clustering will in turn facilitate learning latent representations.

To optimize the DNN parameters $\theta$, $\psi$ and latent representation $\bm{z}$, we observe that in Equation~\ref{eq:all_elbo}, only the first three terms make a contribution. We denote this part in the ELBO as $\mathcal{L}_{\textrm{ELBO-VAE}}$, which is derived by combining the first three terms involving $\bm{z}$ in Equation~\ref{eq:all_elbo} and we optimize $\mathcal{L}_{\textrm{ELBO-VAE}}$ to estimate the DNN parameters and $\bm{z}$ while fixing the DPMM parameters. The notations in Equation~\ref{eq:elbo_derivation} can be found in Table~\ref{tab:notation} and derivation details are provided in Section 1 in the Supplement. 
\begin{align}
&\mathcal{L}_{\textrm{ELBO-VAE}}(\bm{x})\notag=\\ & -\frac{1}{2L}\sum_{l=1}^L \sum_{k=1}^{T} N_k \nu_k  \textrm{Tr}(S_k W_k)\notag\\
&-\frac{1}{2L}\sum_{l=1}^L \sum_{k=1}^{T} N_k \nu_k \left\{ (\bar{\bm{z}}_k-\bm{m}_k)^T W_k (\bar{\bm{z}}_k - \bm{m}_k)\right\} \notag\\
&-\frac{1}{2}\frac{1}{L}\sum_{l=1}^{L}\sum_{i=1}^N\sum_{j=1}^D\left(\log(\bm{\sigma}_x^2)_j^{(l)}+ 
    \frac{\left(\bm{x}_{ij}- (\bm{\mu}_x)_j^{(l)}\right)^2}{(\bm{\sigma}_x^2)_j^{(l)}}\right) \notag\\
    &+\frac{1}{2}\log(\textrm{Det}(2\pi e \Sigma)).
\label{eq:elbo_derivation}
\end{align}

Note that in  Equation~\ref{eq:elbo_derivation}, $\bar{\bm{z}}_k$ is a function of parameters $\theta$ and $\psi$, which are optimized through stochastic gradient descent, where $\bar{\bm{z}}_k=\frac{1}{N_k}\sum_{n=1}^{N}\gamma_{nk}\hat{\bm{z}}_n$ and  $\hat{\bm{z}}_n=\bm{\mu}(\bm{x}_n;\psi)$. In Equation~\ref{eq:elbo_derivation}, the first term comes from training DPMMs on $\bm{z}$. The second term and third term stem from the output from the decoder and encoder respectively, where $\Sigma$ represents the diagonal covariance matrix of the encoder.

\begin{table}[!htb]
  \caption{Notations in the ELBO.} 
   \label{tab:notation}
   \small 
   \centering 
   \begin{tabular}{l} 
   \toprule[\heavyrulewidth]\toprule[\heavyrulewidth]
   \textbf{Notations in the ELBO} \\ 
   \midrule
   $N$: the total number of observations.\\
   $L$: the number of Monte Carlo samples to use in SGVB.\\ 
   $\Sigma$: the diagonal covariance matrix of the encoder. \\
   $\bm{x}_n$: the $n$th observation. \\
   $\bm{y}_n$: cluster membership for the $n$th observation.\\ 
   $p(\bm{y}_n=k) = \gamma_{nk}, N_k = \sum_{n=1}^N \gamma_{nk}$.  \\
   $\hat{\bm{z}}_n=\bm{\mu}(\bm{x}_n;\psi)$: the estimated mean of $\bm{z}_n$ given $\bm{x}_n$. \\
   $\bar{\bm{z}}_k=\frac{1}{N_k}\sum_{n=1}^{N}\gamma_{nk}\hat{\bm{z}}_n$.\\
     $\bm{S}_k = \frac{1}{N_k}\sum_{n=1}^N \gamma_{nk} (\hat{\bm{z}}_n - \bar{\bm{z}}_k)(\hat{\bm{z}}_n - \bar{\bm{z}}_k)^T.$ \\
     $\beta_k = \beta_0+ N_k$: the scalar precision in NW distribution. \\
    $\bm{m}_k=\frac{1}{\beta_k}(\beta_0\bm{m}_0+N_k\bar{z}_k)$: the posterior mean of cluster $k$. \\
    $W_k^{-1} = W_0^{-1} + N_k S_k + \frac{\beta_0 N_k}{\beta_0 + N_k}(\bar{\bm{z}}_k- \bm{m}_0)(\bar{\bm{z}}_k- \bm{m}_0)^T$.\\ 
    $\nu_k = \nu_0+N_k$: the $k$th posterior degrees of freedom of NW. \\
    $\bm{\phi}$: variational parameters of the $k$th NW components.\\
   \bottomrule[\heavyrulewidth] 
   \end{tabular}
\end{table}

\subsection{Streaming Variational Inference}
In this section, we provide an outline for updating the DPMM parameters in a streaming setting, where the latent representation $\bm{z}$ is treated as the observations for DPMMs. We have encountered mainly three challenges including catastrophic forgetting of DNNs, efficient sequential posterior approximation and incremental sufficient statistics updates with sequential data streams. We describe our solution to each challenge below. 

\subsubsection{DGR to Overcome Catastrophic Forgetting}
Catastrophic forgetting of DNNs refers to the fact that DNNs will overfit on new data and discard previously learned information when changes exist in the data distribution, which has been shown in \citet{kirkpatrick2017overcoming, shin2017continual} in a supervised classification setting. We find this is also the case for AdapVAE in an unsupervised setting. This is a dramatic issue for streaming data since if DNNs forget the latent representation for previously learned clusters, when we pass the latent representation to DPMMs, DPMMs will not recover the previously learned clusters either. To resolve the issue, we adopt DGR to generate a fixed small number of samples based on the current DNN and DPMM parameter estimates and merge the samples with new data streams to further update the DNN and DPMM parameters with the estimates in the previous iteration as initialization and the approximation of the posterior as the prior distribution for the next data stream. DGR is a natural choice for our framework since generating a small number of samples is a computationally efficient byproduct of AdaptVAE if memory management of storing historical data is the bottleneck.

\subsubsection{Sequential Bayesian Updating Rule}\label{sec:sbur}
To approximate the posterior via streaming variational inference, we integrate the Bayesian updating rule in our inference by exploiting the sequential nature of Bayes theorem to recursively update an approximation of the posterior and use it as a new prior for incoming data. This is valid since assuming the posterior is $P(\bm{y}, \bm{z}, \bm{\phi}, \bm{v}|\mathcal{D}_1, \mathcal{D}_2, \ldots, \mathcal{D}_{b-1})$ given $(b-1)$ data streams $\mathcal{D}_{i}$, $i=1, 2, \ldots, (b-1)$, the posterior after the $b$th data stream arrives is
\begin{align}&P(\bm{y}, \bm{z}, \bm{\phi}, \bm{v}|\mathcal{D}_1, \mathcal{D}_2, \ldots, \mathcal{D}_{b})\notag\\
&\propto P(\mathcal{D}_b|\bm{y}, \bm{z}, \bm{\phi}, \bm{v})P(\bm{y}, \bm{z}, \bm{\phi}, \bm{v}|\mathcal{D}_1,\ldots, \mathcal{D}_{b-1}).
 \label{eq:bayesian}
\end{align}
That is, the posterior of $(b-1)$ data streams can be considered as the prior for the $b$th data stream. If we know exactly $P(\bm{y}, \bm{z}, \bm{\phi}, \bm{v}|\mathcal{D}_1, \mathcal{D}_2, \ldots, \mathcal{D}_{b-1})$ and the normalizing constant for $P(\bm{y}, \bm{z}, \bm{\phi}, \bm{v}|\mathcal{D}_1, \mathcal{D}_2, \ldots, \mathcal{D}_{b})$, repeatedly updating Equation~\ref{eq:bayesian} is streaming without reusing past data. However, in reality, it is often infeasible to know $P(\bm{y}, \bm{z}, \bm{\phi}, \bm{v}|\mathcal{D}_1, \mathcal{D}_2, \ldots, \mathcal{D}_{b-1})$ and the normalizing constant. Thus, we adopt approximation of the posterior via variational inference. 

\subsubsection{Incremental Updates using Hierarchical Sufficient Statistics}
The third challenge for streaming data is to develop an efficient incremental algorithm that can optimize the objective function using summary statistics of the encountered data streams instead of the entire dataset. Our algorithm performs incremental updates using different levels of summary statistics for the encountered data streams. Assuming we have encountered $b$ data streams $\{\mathcal{D}_j\}_{j=1}^{b}$, we divide the data stream $\mathcal{D}_j$ in memory  into $M$ mini-batches $\{\mathcal{B}_i\}_{i=1}^M$. 
We define the global parameters of DPMMs as the stick-breaking proportions $\pi_{k}s$ and data generating parameters $(\bm{\mu}_k^*, \bm{\sigma}_k^*)$ and the local parameters as the cluster assignment $\bm{y}_n$ of $\bm{z}_n$. 
Our algorithm memorizes
three levels of sufficient statistics including the mini-batch sufficient statistics $S_k^{j,i}=(N_k(\mathcal{B}_i), s_k(\mathcal{B}_i))$ of a mini-batch $\mathcal{B}_i$ of  $\mathcal{D}_j$ in memory, where $s_k(\mathcal{B}_i)=\sum_{n\in\mathcal{B}_i}\hat{\gamma}_{nk}t(z_n)$, the stream sufficient statistics $S_k^j=\sum_{i=1}^M
S_k^{j,i}$ of  $\mathcal{D}_j$ and the overall sufficient statistics $S_k^0 = (N_k, s_k(\bm{z}))$ of $\{\mathcal{D}_j\}_{j=1}^{b}$. Given data stream $\mathcal{D}_j$, for each iteration, we subtract the old  summary of $\mathcal{B}_i$, update local parameters $\bm{y}_n$ for each $\mathcal{B}_i$, compute a new summary for $\mathcal{B}_i$, and then update the stream sufficient statistics for each cluster by performing
\begin{align}
&S_k^j \leftarrow  S_k^j - S_k^{j,i},   \\
&S_k^{j, i} \leftarrow\left(\sum_{n\in\mathcal{B}_i} \hat{\gamma}_{nk}, \sum_{n\in\mathcal{B}_i} \hat{\gamma}_{nk} t(\bm{z}_n) \right), \\
&S_k^j \leftarrow  S_k^j + S_k^{j,i}.  
\end{align}
We repeat the process for multiple iterations for $\mathcal{D}_j$, where $t(\bm{z}_n)$ is the sufficient statistics to represent a distribution within the exponential family and $\hat{\gamma}_{nk}$ represents the estimated probability of the $n$th observations in mini-batch $\mathcal{B}_i$ assigned to cluster $k$. Finally, we update the overall sufficient statistics by
$ S_k^0\leftarrow S_k^0 + S_k^j.$
The correctness of the algorithm is guaranteed by the additive property of the sufficient statistics. Our algorithm is different from \citet{hughes2013memoized} since their work is designed for batch mode. Hence, they can replicate multiple passes of the entire dataset. Our algorithm is streaming, which means that we only have one pass for all data streams. However, we can have multiple passes for each mini-batch of the current data stream. After the current data stream is no longer in memory, the approximation of the posterior is considered as the prior for the next data stream as described in Section~\ref{sec:sbur}. We take advantages of birth and merge moves to produce novel clusters or combine them to remove redundancy. The details of  performing birth and merge moves can be found in the Supplement. We summarize the streaming inference procedure in Algorithm~\ref{alg:algo}. Code will be made available on github upon publication.


\begin{algorithm}[!htb]
\caption{\ \\
\textcolor{white}{===} AdapVAE Streaming Variational Inference }
\label{alg:algo}
\begin{algorithmic}[1]
{
\small
\STATE \textbf{Initialization:} \\
DNN parameters, variational distributions and the hyper-parameters of DPMMs. \\

\FOR{the $b$th data stream $\mathcal{D}_b$, $b=1, 2, \ldots$}
\STATE Generate  samples using the current AdapVAE and merge it with $\mathcal{D}_b$.

\STATE Divide the merged $\mathcal{D}_b$ into M mini-batches.
\FOR{the $i$th mini-batch $\mathcal{B}_i$, $i=1, 2, \ldots, M $}

\FOR{$t=1, 2, \ldots, T $}
\STATE Update DNN parameters of AdapVAE by maximizing $ L_{\textrm{ELBO-VAE}}(x)$ in Equation~\ref{eq:elbo_derivation} given current DPMM parameters.
\ENDFOR
 \STATE Compute the latent  $\bm{z}$ of $\mathcal{B}_i$.
\STATE Assign the current approximation of the posterior as the prior for new data and query the cached stream sufficient statistics $S_k^j$. 
\WHILE{The ELBO has not converged}
\STATE Update the local and global DPMM parameters, mini-batch and stream sufficient statistics $S_k^{j,i}$ and $S_k^j$ by Equation~$5$-$7$.
\STATE Birth moves to create new candidate clusters.
\STATE Merge moves to combine similar clusters if the ELBO improves.
\STATE Recompute the ELBO. 
 \ENDWHILE
\STATE Update the DPMM and cache the global DPMM parameters and the overall sufficient statistics.
\ENDFOR
\ENDFOR
}
\STATE \textbf{Output}: Learned AdapVAE, variational distributions.
\end{algorithmic}
\end{algorithm}
\section{Experiments}\label{sec:experiment}
In this section, we want to investigate: (1) how sensitive our algorithm is to detect novel clusters in the new data given different numbers of samples from the new cluster; (2) whether incorporating DGR in AdapVAE in a streaming setting can reproduce cumulative input distributions of all encountered data streams without forgetting what has been learned in the past and capable of detecting novel clusters; and (3) whether the clustering performance of AdapVAE in a batch mode is comparable to state-of-the-art methods (VaDE \citep{jiang2017variational} and DEC \citep{xie2016unsupervised}) using multiple clustering quality evaluation metrics.

\subsection{Datasets}
We evaluate our method using both text and image benchmark datasets, which have been used by \citet{jiang2017variational, xie2016unsupervised}. We mainly use MNIST \citep{lecun1998gradient} to interpret our results and a detailed description of MNIST database is provided. Descriptions for more complex text datasets REUTERS-10k and higher dimensional image dataset STL-10 are provided in the supplement. 
MNIST consists of images of 70,000 handwritten digits of $28\times28$ pixel size. In order to compare fairly with previous methods, we normalized and flattened each image to a $784\times1$ vector and adopted the same neural network architecture as DEC and VaDE. The pipeline is $d-500-500-2000-l$ and $l-2000-500-500-d$ for the encoder and decoder, respectively, where $d$ and $l$ denote the dimensionality of the input and latent features. All layers are fully connected and a sampling layer bridges the encoder and decoder. We adopt the same pre-trained Stacked Autoencoder in VaDE as the initialization for the neural network. Adam optimizer \citep{kingma2014adam} is used as the optimization engine to update the neural network. The batch size is set to 1500 for a batch setting. The size of each data stream is set to 1000 and we divide it into two mini-batches. The learning rate for Reuters-10K and STL-10 is set as 0.002 and 0.0002 for MNIST with a common decay rate of 0.9 for every epoch in a batch setting. For the clustering initialization, DEC and VaDE start with K-means or GMM and fix the number of clusters as the ground truth. AdapVAE starts with one cluster and grows or merges clusters using birth and merge moves.

Since DPMMs may over-cluster the digits into more than six clusters, we use Normalized Mutual Information (NMI), Adjusted Rand Index (ARI), Homogeneity Score (HS) and V-measure Score (VM) to compare clustering quality for results with different numbers of clusters. They are all normalized metrics ranging from zero to one with value one representing perfect clustering as the ground truth. Detailed definition of each metric is provided in the Supplement.


\subsection{Sensitivity of Novel Cluster Detection}
We choose MNIST  to explore how sensitive our algorithm is to detect novel clusters in the new data. We train AdapVAE using a random sample of 10,000 images from digits 0-4 and then randomly choose another total number of 10,000 samples for all digits 0-5 of MNIST with equal number of samples per digit and divide them into 10 sequential data streams of equal size. We vary the percentage of new digit 5 samples from 1\%, 2\%, 5\%, 10\%, 15\% and 20\% out of 10000 samples and examine the precision and recall of novel cluster detection and the overall clustering performance. Digit 5 is chosen as the new cluster since we show in Section 6.3 that it is the most challenging digit for novel cluster detection with the lowest recall.  

\begin{figure}[h]
   \centering
	\includegraphics[scale=0.25]{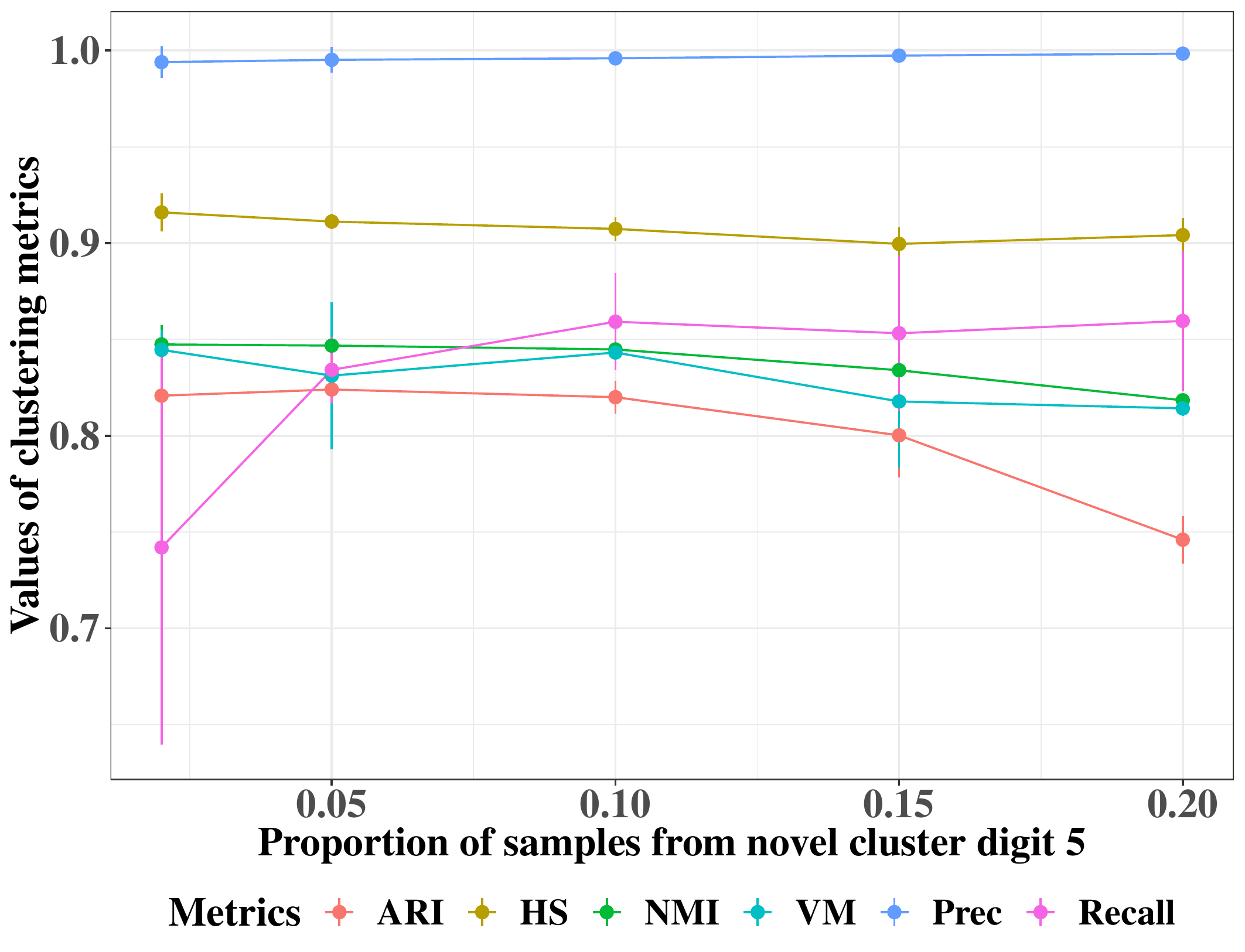}
	\caption{Clustering metric values with different proportions of new cluster samples out of the total across five replications.}
    \label{fig:five}
\end{figure}

\begin{figure}[h]
   \centering
	\includegraphics[scale=0.2]{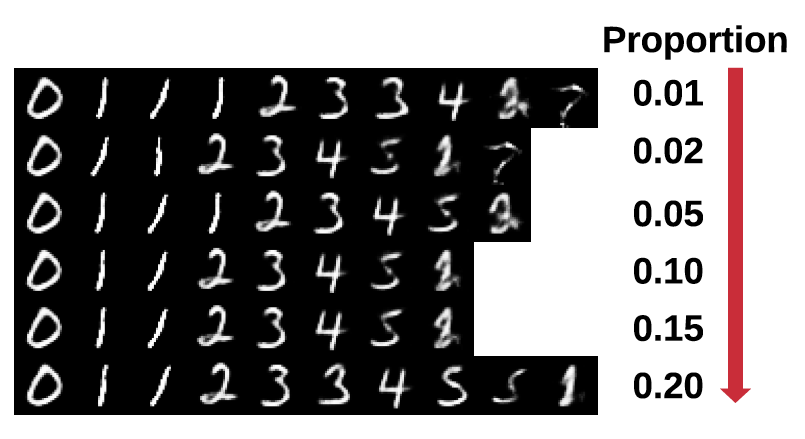}
	\caption{Decoded images using the DPMM posterior mean with different proportions of digit 5 samples.}
    \label{fig:prop}
\end{figure}

We report the values for clustering metrics in Figure~\ref{fig:five} and the decoded images using the learned DPMM posterior mean with different proportions of new digit 5 samples in Figure~\ref{fig:prop}.

Figure~\ref{fig:prop} reflects that the algorithm is not able to detect the new cluster with only 1\% of digit 5 samples out of the total.
When the proportion increases to 2\%, the algorithm is able to group 76.1\% of all digit 5 samples into a new cluster. The recall of novel cluster detection increases as the percentage increases. When the percentage increases to 20\%, AdapVAE clusters the digit 5 samples into two new sub-clusters with different handwritten styles. We merge the two subclusters into one and report the overall precision and recall when the percentage is 20\%. As the proportion increases, metrics such as ARI and HS decrease since AdapVAE groups digit 5 samples into two sub-clusters while the ground truth expects all digit 5 samples to form one cluster. The decoded image quality of digit 5 increases as the proportion increases. We  observe that the last image of each row represents a mixture clusters with samples from all digits. In summary, AdapVAE has good sensitivity for novel cluster detection. 

\subsection{Overcoming Catastrophic Forgetting and Online Novelty Detection}
We demonstrate that integrating DGR in AdapVAE can help recover information learned through past data. We divide the MNIST dataset into 5 disjoint subsets and each data stream contains 10,000 random samples of two digit clusters in the order of digits 0-1, 2-3, 4-5, 6-7 and 8-9 respectively, denoted as DS1, DS2, DS3, DS4, and DS5. AdapVAE can only get access to one data stream each time. The mini-batch size within each data stream is 500. We generate 100 samples using DGR for each mini-batch. We sequentially train AdapVAE with data streams DS1, DS2, DS3, DS4, and DS5. This experiment setup is challenging since there are no samples in the current data stream that are from clusters learned in previous data.

We report the precision and recall of detecting the novel clusters across five replications in Table~\ref{tb:novelty}. We provide the decoded image using posterior DPMM mean  parameters and the weights of the decoder in AdapVAE, which is sequentially updated using DS1, DS2, DS3, DS4, and DS5. 

Without DGR, we find that after training with digits 2 and 3, AdapVAE will decode the posterior mean of digit 0 and 1 in the latent space learned previously into digits 2 and 3 in the original data space, which indicates the DNNs have forgotten the information learned with training data of digits 0 and 1. 

\begin{figure}[h]
   \centering
  
	\includegraphics[scale=0.46]{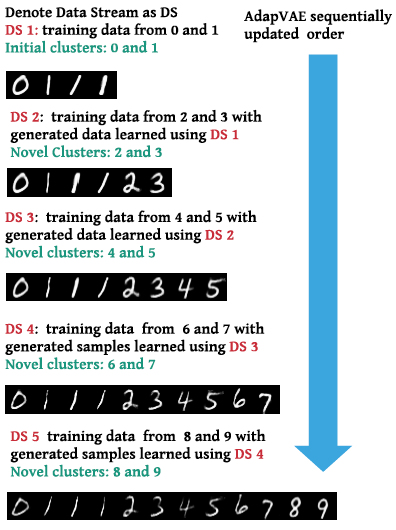}
	\caption{ We plot the decoded image of the posterior DPMM mean parameters using AdapVAE, which is sequentially updated using data streams each with two digits incrementally.}
    \label{fig:novel}
\end{figure}

Figure~\ref{fig:novel} reflects that AdapVAE with DGR is able to remember and reproduce cumulative input distributions of all encountered data streams and is able to generate samples from all learned clusters with examples shown in the Supplement. Table~\ref{tb:novelty} reflects that AdapVAE is able to detect novel clusters in new data with high precision and recall if trained sequentially with mutually exclusive subsets of clusters in each data stream. We also find that the new digit clusters get immediately assigned to a new cluster in DPMMs through birth moves while training.
The precision for digits 8 and 9 is relatively low with large deviation since AdapVAE merges samples from digit 4 and digit 5 with digits 9 and 8 respectively. Detecting digit 5 is the most challenging case since it has the lowest recall of 77.45\%, which indicates 22.55\% of the new samples from digit 5 gets assigned to previously learned digit clusters of digits 0-4.

\begin{table}[h]
\caption{Mean (standard error) of precision and recall for each two novel clusters in each sequential data stream across five replications in percentage.}
\label{tb:novelty}
\small
\begin{center}
 \begin{tabular}{||c| c c ||} 
 \hline
Novel Clusters & Precision & Recall \\
 \hline\hline
0& 100 (0) & 96.60 (3.77)  \\ 
1 (sub-cluster1) & 100 (0)& 50.46 (1.63)\\
1 (sub-cluster2) &100 (0)& 37.13 (1.96)\\
1 (sub-cluster3) &81.68 (19.3)& 10.34 (3.35)\\
\hline\hline
2& 92.96 (3.91) & 99.64 (0.56) \\
3&99.21 (0.89) & 95.20 (2.48) \\\hline\hline
4& 98.80 (0.68) & 99.62 (0.52) \\
5& 99.64 (0.69) & \textbf{77.45} (\textbf{5.36}) \\\hline\hline
6& 99.64 (0.50) & 97.03 (1.75) \\
7&97.52 (2.75)& 97.80 (3.35)  \\\hline\hline
8& \textbf{88.03} (\textbf{9.70}) & 98.57 (1.50)  \\
9&\textbf{83.47} (\textbf{15.0})& 96.29 (1.54) \\\hline\hline
\end{tabular}
\end{center}

\end{table}

\subsection{Clustering Performance Comparison given Data in a Batch Setting}
Since there is no comparable streaming algorithm that can automatically learn features using DNNs and simultaneously cluster in a nonparametric setting, we compare the clustering performance of AdapVAE in a batch mode with DEC, VaDE and VAE plus DP learned separately, where VaDE and DEC have knowledge about the ground truth number of clusters.

\begin{table}[!htb]
\caption{Clustering quality (\%) comparison averaged over five replications with both the average value and the standard error (in the parenthesis) provided.}
\label{tb:multiple}
\begin{center}
\small

\begin{tabular}{c | c c c }
    \hline
    Dataset &  Method & NMI & ARI \\ [0.5ex]  
 \hline\hline
\multirow{3}{*}{MNIST}& DEC & 84.67 (2.25) & \textbf{83.67} (4.53) \\
                        &VaDE &   80.35 (4.68) & 74.06 (9.11)  \\
                       & VAE+DP & 81.70 (0.825) & 70.49 (1.654)  \\
                        &AdapVAE & \textbf{85.72} (\textbf{1.02})& 83.53 (\textbf{2.35})\\
\hline
\multirow{3}{*}{Reuters10k}& DEC & \textbf{46.56} (5.36) & \textbf{46.86} (7.98) \\ 
& VaDE & 41.64 (4.73) & 38.49 (\textbf{5.44}) \\
& VAE + DP & 41.62 (2.99) & 37.93 (4.57) \\
& AdapVAE & 45.32 (\textbf{1.79}) & 42.66 (5.73)  \\
\hline
\multirow{3}{*}{STL10} & DEC & 71.92 (2.66) & 58.73 (5.09) \\
& VaDE & 68.35 (3.85) & 59.42 (6.84) \\
&VAE+DP & 43.18 (1.41) & 26.58 (1.32)  \\
&AdapVAE &\textbf{75.26} (\textbf{0.53}) & \textbf{70.72} (\textbf{0.81}) \\
\hline

    Dataset &  Method & HS & VM \\ [0.5ex]  
 \hline\hline
\multirow{3}{*}{MNIST}& DEC  & 84.67 (2.25) & 84.67 (2.25)   \\
                        &VaDE & 79.86 (4.93) & 80.36 (4.69)  \\
                       & VAE+DP  & \textbf{91.27} (\textbf{0.215}) & 81.19 (0.904)  \\
                        &AdapVAE & 89.34 (0.25)  & \textbf{85.65} (\textbf{0.51}) \\
\hline
\multirow{3}{*}{Reuters10k}& DEC & 48.44 (5.44) & \textbf{46.52} (5.36)  \\ 
& VaDE &  43.64 (4.88) & 41.60 (4.73)\\
& VAE + DP & 46.64 (3.85) & 41.34 (2.94) \\
& AdapVAE & \textbf{48.88} (\textbf{1.86})  & 45.40 (\textbf{2.04})\\
\hline
\multirow{3}{*}{STL10} & DEC  & 68.47 (3.48) & 71.83 (2.72) \\
& VaDE & 67.24 (4.23) & 68.37 (3.92) \\
&VAE+DP  & 42.28 (1.03) & 43.16 (1.39) \\
&AdapVAE & \textbf{77.61} (\textbf{1.29})  & \textbf{75.22} (\textbf{0.52}) \\
\hline
\end{tabular}
\end{center}
\end{table}

The performance of AdapVAE and DEC is more stable. Training VAE and DP separately as VAE+DP will over-cluster the digits in MNIST into 19-26 clusters. Thus, it has the best homogeneity for MNIST. AdapVAE finds between 11 and 15 clusters and the result usually contains one mixture cluster with samples from all digits for cases when the handwritten digits are not clear to tell. AdapVAE clusters some digits into multiple sub-clusters with different writing styles such as upright and oblique digit one. The best performance of VaDE out of five replications is comparable to other methods but it has larger variation. 

\section{Conclusion}
We presented a novel clustering algorithm AdapVAE combining DPMM prior with VAEs for streaming data. It provides an end-to-end deep representation of the data in a low-dimensional latent space with rich clustering structure. We develop a streaming variational inference algorithm to update both the neural network and DPMM parameters. Our work can adaptively detect novel clusters in an online fashion. Both qualitative and quantitative analysis for both text and image benchmarks are provided. 

\section{Acknowledgements}
We would like to acknowledge support for this project from NIH/NHLBI U01HL089856 and NIH/NCI R01CA199673.




\section{SUPPLEMENTARY MATERIAL}


\subsection{Variational Inference for AdapVAE and ELBO Derivation} 
In this section, we provide the ELBO derivation. Recall that we use the variational distribution $q(\bm{y}, \bm{z}, \bm{\phi}, \bm{v}|\bm{x})$ to approximate the posterior distribution $p(\bm{y}, \bm{z}, \bm{\phi}, \bm{v}|\bm{x})$. Minimizing the Kullback-Leibler (KL) divergence between $q(\bm{y}, \bm{z}, \bm{\phi}, \bm{v}|\bm{x})$ and $p(\bm{y}, \bm{z}, \bm{\phi}, \bm{v}|\bm{x})$ is equivalent to maximizing the ELBO $\mathcal{L}_{\textrm{ELBO}}$. We first list the assumptions on the variational distribution $q(\bm{y}, \bm{z}, \bm{\phi}, \bm{v}|\bm{x})$ and then provide the ELBO derivation and the updating equations. 

We assume that 
\begin{align*}
  q(\bm{y}, \bm{z}, \bm{\phi}, \bm{v}|\bm{x}) &= q_{\psi}(\bm{z}|\bm{x})q(\bm{y, z, \phi}|\bm{x}) \\
  &=q_{\psi}(\bm{z}|\bm{x})q(\bm{y})q(\bm{v})q(\bm{\phi}).
\end{align*}

Now, we list the variational distribution assumptions for $q_{\psi}(\bm{z}|\bm{x})$, $q(\bm{y})$, $q(\bm{v})$ and $q(\bm{\phi})$
respectively.
\begin{align}
q(\bm{y, \bm{v}, \phi}|\bm{x}) = \prod_{t=1}^{T-1} q_{\eta_t}(v_t)\prod_{t=1}^T q_{\zeta_t}(\bm{\phi}_t)\prod_{n=1}^{N}q_{\rho_n}(y_n),
\label{eq:variational}
\end{align} 
where $T$ is the number of mixture components in the DP of the variational distribution and $\bm{z}_n\sim \mathcal{N}(\bm{z}_n|\mu_t, \Lambda_t^{-1}).$
\begin{itemize}
    \item $q_{\psi}(\bm{z}|\bm{x}) =  \mathcal{N}(\bm{\mu}(\bm{x};\psi), \bm{\sigma}^2(\bm{x}; \psi))$
    \item $q_{\zeta_t}(\phi_t)=q(\mu_t|\Lambda_t)q(\Lambda_t)$, $q(\mu_t|\Lambda_t)q(\Lambda_t)=\mathcal{N}(\mu_t|m_t, (\beta_t\Lambda_t)^{-1})\mathcal{W}(\Lambda_t|W_t, \nu_t)$, where $\phi_t=(\mu_t, \Lambda_t)$.
    \item $q(y_n) = \textrm{Mult}(T, \rho_n)$, which is a Multinomial
    distribution.
    \item $q_{\eta_t}(v_t) = \textrm{Beta}(\eta_{t_1}, \eta_{t_2})$.
\end{itemize}
Under our assumptions, the $\mathcal{L}_{\textrm{ELBO}}(x)$ can be rewritten as:
 \begin{align} \label{first}
            \begin{split}
&\mathcal{L}_{\textrm{ELBO}}(x) = \mathbb{E}_{q(\bm{y}, \bm{z}, \bm{\phi}, \bm{v}|\bm{x})}\left[
    \log\frac{p(\bm{x}, \bm{y}, \bm{z}, \bm{\phi}, \bm{v})}{q(\bm{y}, \bm{z}, \bm{\phi}, \bm{v}|\bm{x})}\right]
    \\
    &=\mathbb{E}_{q(\bm{y}, \bm{z}, \bm{\phi}, \bm{v}|\bm{x})}
    \left[\log\frac{p(\bm{x}|\bm{z})p(\bm{z}|\bm{y}, \bm{\phi})p(\bm{y}|\bm{v})p(\bm{v})p(\bm{\phi})}{q_{\psi}(\bm{z}|\bm{x})q(\bm{y}|\bm{x})q(\bm{v})q(\bm{\phi}|\bm{x}))}\right]
    \\
    &= \mathbb{E}_{q(\bm{y}, \bm{z}, \bm{\phi}, \bm{v}|\bm{x})}
    \left[\log p(\bm{x}|\bm{z})\right]
    +\mathbb{E}_{q(\bm{y}, \bm{z}, \bm{\phi}, \bm{v}|\bm{x})}
    \left[\log p(\bm{z}|\bm{y}, \bm{\phi})\right]
    \\
    &+ \mathbb{E}_{q(\bm{y}, \bm{z}, \bm{\phi}, \bm{v}|\bm{x})}
    \left[\log p(\bm{y}|\bm{v})\right] 
    + \mathbb{E}_{q(\bm{y}, \bm{z}, \bm{\phi}, \bm{v}|\bm{x})}
    \left[\log p(\bm{v})\right] 
    \\
    &+ \mathbb{E}_{q(\bm{y}, \bm{z}, \bm{\phi}, \bm{v}|\bm{x})}
    \left[\log p(\bm{\phi})\right] 
    - \mathbb{E}_{q(\bm{y}, \bm{z}, \bm{\phi}, \bm{v}|\bm{x})}
    \left[\log q_{\psi}(\bm{z}|\bm{x})\right] 
    \\
    &- \mathbb{E}_{q(\bm{y}, \bm{z}, \bm{\phi}, \bm{v}|\bm{x})}
    \left[\log q(\bm{y}|\bm{x})\right] 
    - \mathbb{E}_{q(\bm{y}, \bm{z}, \bm{\phi}, \bm{v}|\bm{x})}
    \left[\log q(\bm{v})\right] 
    \\
    &- \mathbb{E}_{q(\bm{y}, \bm{z}, \bm{\phi}, \bm{v}|\bm{x})}
    \left[\log q(\bm{\phi}|\bm{x})\right] 
        \end{split}
        \end{align}
In our updating strategy, we adopt an alternating optimization strategy used by \citet{goyal2017nonparametric}. To be specific, we update the VAE parameters ($\theta$ and $\psi$) and the latent variable ($\bm{z}$) given the current estimates of the DPMM parameters. When updating the DPMM parameters, the latent representation $\bm{z}$ is treated as the observations for DPMM. The updates for the local cluster membership assignment parameter $\bm{y}$, global parameters $\bm{\phi}$ and $\bm{v}$ simplifies to the updates in variational inference for DPMM developed by \citet{blei2006variational}. Hence, we only list the updating equations and the expectation derivation involving $q(\bm{y})$, $q(\bm{\phi})$ and $q(\bm{v})$ at the end of this section. 

The notation summary is provided in the main paper. We focus on deriving the nonstandard terms involving the VAE parameters $\theta$, $\psi$ and latent representation $\bm{z}$.

(1) $\mathbb{E}_{q_{\psi}(\bm{z}|\bm{x})q(\bm{y})q(\bm{v})q(\bm{\phi})}\left[\log P_\theta(\bm{x}|\bm{z})\right]$:

We use  a neural network $g$ to model the decoder with parameters $\theta$, where $
       (\bm{\mu_x}, \log\bm{\sigma_x}^2)=g_\theta(\bm{z})$ and $ 
        P_\theta(\bm{x}|\bm{z}) = \mathcal{N}(\bm{x};\bm{\mu_x},\bm{\sigma_x}^2\bm{I}).
       $ Hence, we have
       
\begin{align} 
  \mathbb{E}&_{q_{\psi}(\bm{z}|\bm{x})q(\bm{y})q(\bm{v})q(\bm{\phi})}[\log P_\theta(\bm{x}|\bm{z})]\notag\\
    &=\mathbb{E}_{q_{\psi}(\bm{z}|\bm{x})q(\bm{y})q(\bm{v})q(\bm{\phi})}\left(-\frac{1}{2}\log(\bm{\sigma_x}^2)_j +\frac{(\bm{x}_j-(\bm{\mu_x})_j)^2}{(\bm{\sigma_x}^2)_j}    \right)
    \notag\\
    &=-\frac{1}{2}\frac{1}{L}\sum_{i=1}^N\sum_{j=1}^D\left(\log(\bm{\sigma}_x^2)_j^{(l)}+ 
    \frac{\left(\bm{x}_{ij}- (\bm{\mu}_x)_j^{(l)}\right)^2}{(\bm{\sigma}_x^2)_j^{(l)}}\right),
        \label{eq:3}
\end{align}
where
\begin{itemize}
    \item $(\bm{\mu_x})_j$: represents the $j$th element of $\bm{\mu_x}$ for $j=1,2, \dots, D$.
\item
$(\bm{\sigma_x}^2)_j$: represents the $j$th element of $\bm{\sigma_x}^2$ for $j=1,2, \ldots, D$.
\item
  $\bm{x}_{ij}$: represents the $j$th element of the $i$th observation.
\end{itemize}

(2) $\mathbb{E}_{q_{\psi}(\bm{z}|\bm{x})q(\bm{y})q(\bm{v})q(\bm{\phi})}[\log p(\bm{z}|\bm{y}, \bm{\phi})]:$ 

We use neural network $f$ to model the encoder with parameters $\psi$,
where $q_\psi(\bm{z}|\bm{x})= \mathcal{N}(\bm{\mu}(\bm{x};\psi), \bm{\sigma}^2(\bm{x}; \psi))$ and $
        (\bm{\mu}(\bm{x};\psi), \log\bm{\sigma}^2(\bm{x}; \psi))=f(\bm{x}; \psi).$ In VAE, we use the reparameterization trick to allow backpropagation:
\\
\begin{equation*}
\bm\epsilon^{(l)}\sim\mathcal{N}(\bm{0}, \bm{I}) ~\ \textrm{and} ~\ \bm{z}^{(l)}= \bm{\mu}(\bm{x};\psi)+\bm{\epsilon}^{(l)}\bm{\sigma}(\bm{x}; \psi).
\end{equation*}
We denote $ \hat{\bm{z}}_n$ as the estimated mean of the latent representation from the encoder given $x_n$:
\begin{equation*}
    \hat{\bm{z}}_n = g_{\mu}(\bm{x}_n; \phi)=\bm{\mu}(\bm{x}_n;\psi) = \underset{L\rightarrow \infty}{\mbox{lim}}\frac{1}{L}\sum_{l=1}^L \bm{z}_n^{(l)}.
\end{equation*}
According to Equation 10.71 of \citet{bishop2006pattern}, we have the following:

\begin{align} 
\begin{split}
&\mathbb{E}_{q_{\psi}(\bm{z}|\bm{x})q(\bm{y})q(\bm{v})q(\bm{\phi})}[\log p(\bm{z}|\bm{y}, \bm{\phi})]\notag
\\ &=\frac{1}{2}\sum_{k=1}^T N_k\left\{
\log \tilde{\Lambda}_k - D\beta_k^{-1} - \nu_k\mbox{Tr}(S_k W_k)\right\}\notag\notag\\
&-\frac{1}{2}\sum_{k=1}^T N_k\left\{\nu_k(\bar{z}_k -\bm{m}_k)^T W_k(\bar{z}_k -\bm{m}_k)- D\log(2\pi)
\right\},
\end{split}
\end{align} 
where
\begin{align} 
  \label{third}
  \begin{split}
    &\mathbb{E}_{\phi_k}\left[\left(\hat{\bm{z}}_n -\mu_k\right)^T\Lambda_k\left(\hat{\bm{z}}_n -\mu_k\right)\right]\notag\\
    &=\frac{D}{\beta_k} + \nu_k\left(\hat{\bm{z}}_n-\bm{m}_k\right)^T W_k\left(\hat{\bm{z}}_n-\bm{m}_k\right)\\
    &
   \log\tilde{\Lambda}_k = \mathbb{E}[\log\Lambda_k]\\
   &= \sum_{j=1}^D\psi\left(\frac{\nu_k+1-i}{2}\right) + D\log 2 + \log\vert W_k\vert.
  \end{split}
\end{align}

(3) $\mathbb{E}{q_{\psi}(\bm{z}|\bm{x})q(\bm{y})q(\bm{v})q(\bm{\phi})}(\log q_{\psi}(\bm{z}|\bm{x}))$:

We assume that $q_\psi(\bm{z}|\bm{x})=  \mathcal{N}(\bm{\mu}(\bm{x};\psi), \bm{\sigma}^2(\bm{x}; \psi))$. Hence, $\mathbb{E}{q_{\psi}(\bm{z}|\bm{x})q(\bm{y})q(\bm{v})q(\bm{\phi})}(\log q_{\psi}(\bm{z}|\bm{x}))$ is equal to the negative entropy of a multivariate Gaussian distribution, which is:
\begin{align} 
            \begin{split}
&\mathbb{E}{q_{\psi}(\bm{z}|\bm{x})q(\bm{y})q(\bm{v})q(\bm{\phi})}(\log q_{\psi}(\bm{z}|\bm{x})) \notag\\
&= \frac{1}{2}\log(\textrm{Det}(2\pi e \Sigma))
    \end{split}
\end{align}
where $\Sigma = \textrm{diag}(\bm{\sigma}^2(\bm{x};\psi))$.

When we update the VAE parameters $\theta$ and $\psi$ and the latent representation $\bm{z}$, the DPMM parameters will be fixed. Thus,
the terms that do not involve $\bm{z}$, $\theta$, $\psi$ will not contribute to the $L_{\textrm{ELBO}-\textrm{VAE}}$. Hence, we obtain
\begin{align}
&\mathcal{L}_{\textrm{ELBO}-\textrm{VAE}}(\bm{x}) \\
&= -\frac{1}{2L}\sum_{l=1}^L \sum_{k=1}^{T} N_k \nu_k \left\{ \textrm{Tr}(S_k W_k)\right\}\notag\\
&-\frac{1}{2L}\sum_{l=1}^L \sum_{k=1}^{T} N_k \nu_k\left\{(\bar{z}_k-\bm{m}_k)^T W_k (\bar{z}_k - \bm{m}_k)\right\} \notag\\
&-\frac{1}{2}\frac{1}{L}\sum_{i=1}^N\sum_{j=1}^D\left(\log(\bm{\sigma}_x^2)_j^{(l)}+ 
    \left(\bm{x}_{ij}- (\bm{\mu}_x)_j^{(l)}\right)^2/(\bm{\sigma}_x^2)_j^{(l)}\right)\notag\\ &+\frac{1}{2}\log(\textrm{Det}(2\pi e \Sigma)).
\label{eq:elbo_derivation}
\end{align}

Here, we list the standard variational inference updating equations and derivations for DPMM.

\begin{itemize}
    \item $q(\bm{y}_n=i)=\gamma_{n, i}.$ 
    \item $q(\bm{y}_n>i)= \sum_{j=i+1}^{T}\gamma_{n, j}.$
    \item $\mathbb{E}_{q}[\log V_i]=\Psi(\gamma_{i, 1})-\Psi(\gamma_{i, 1} + \gamma_{i, 2}).$
    \item $\mathbb{E}_{q}[\log(1-V_i)]=\Psi(\gamma_{i, 2}) - \Psi(\gamma_{i, 1} + \gamma_{i, 2}).$
    \item $\gamma_{n, t}\propto \exp(S_t)$,
    \item $S_t = \mathbb{E}[\log V_i] + \sum_{i=1}^{t-1}\mathbb{E}_q[\log(1-V_i)]+
    \frac{1}{2}\log\tilde{\Lambda}_k-\frac{D}{2\beta_k}-\frac{\nu_k}{2}(\hat{z}_n-\bm{m}_k)^T W_k (\hat{z}_n-\bm{m}_k).$
    \item $\gamma_{n, t} = \frac{exp(S_t)}{\sum_{t=1}^T exp(S_t)}$.
    \item Under the Gaussian-Wishart distribution assumption, \begin{align*} \label{six}
\begin{split}
&\mathbb{E}{q_{\psi}(\bm{z}|\bm{x})q(\bm{y})q(\bm{v})q(\bm{\phi})}(\log p(\bm{\phi})) 
    \\
&=
    \frac{1}{2}\sum_{k=1}^T \left\{
        D\log(\beta_0/2\pi) + \log \tilde{\Lambda}_k\right\}\notag\\
&-\frac{1}{2}\sum_{k=1}^T\left\{\frac{D\beta_0}{\beta_k} +\beta_0\nu_k(\bm{m}_k-\bm{m}_0)^T W_k (\bm{m}_k-\bm{m}_0)\right\} \\
        &+ T \log\mbox{B}(W_0, \nu_0)  \frac{\nu_0-D-1}{2}\sum_{k=1}^T \log\tilde{\Lambda}_k\notag\\ 
&-\frac{1}{2}\sum_{i=1}^T \nu_k\mbox{Tr}(W_0^{-1} W_k),
    \end{split}
\end{align*}
        where 
\begin{align*}
&\mbox{B}(W, \nu) \notag\\
&= \vert W\vert ^{-\nu/2} \left(2 ^{\nu D/2}\pi^{D(D-1)/4}\prod_{i=1}^{D}\Gamma\left(\frac{\nu+1-i}{2}\right)   \right)^{-1}.
\end{align*}
  \item Similarly, we have
  \begin{align*}
            \begin{split}
  & \mathbb{E}{q_{\psi}(\bm{z}|\bm{x})q(\bm{y})q(\bm{v})q(\bm{\phi})}[\log q(\bm{\phi})]\\
   &=
    \sum_{k=1}^T \left(  
        \frac{1}{2}\log \tilde{\Lambda}_k +\frac{D}{2}\log\left(\frac{\beta_k}{2\pi}\right)-\frac{D}{2}- H[q(\Lambda_k)]
    \right),
   \end{split}
\end{align*}

\begin{equation*}
    H[\Lambda] = -\log B(W, \nu)-\frac{\nu-D-1}{2}
    \mathbb{E}[\log |\Lambda|]+ 
   \frac{\nu D}{2}.
\end{equation*}
\end{itemize}

A summary of notations for deriving the ELBO is listed below.

\begin{table}[!htb]
  \caption{Notations in the ELBO.} 
   \label{tab:notation}
   \small 
   \centering 
   \begin{tabular}{l} 
   \toprule[\heavyrulewidth]\toprule[\heavyrulewidth]
   \textbf{Notations in the ELBO} \\ 
   \midrule
      $N$: the total number of observations.\\
   $L$: the number of Monte Carlo samples to use in SGVB.\\ 
   $\Sigma$: the diagonal covariance matrix of the encoder. \\
   $\bm{x}_n$: the $n$th observation. \\
   $\bm{y}_n$: cluster membership for the $n$th observation.\\ 
   $p(\bm{y}_n=k) = \gamma_{ik}, N_k = \sum_{n=1}^N \gamma_{nk}$.  \\
   $\hat{\bm{z}}_n=\bm{\mu}(\bm{x}_n;\psi)$: the estimated mean of $\bm{z}_n$ given $\bm{x}_n$. \\
   $\bar{\bm{z}}_k=\frac{1}{N_k}\sum_{n=1}^{N}\gamma_{nk}\hat{\bm{z}}_n$.\\
     $\bm{S}_k = \frac{1}{N_k}\sum_{n=1}^N \gamma_{nk} (\hat{\bm{z}}_n - \bar{\bm{z}}_k)(\hat{\bm{z}}_n - \bar{\bm{z}}_k)^T.$ \\
     $\beta_k = \beta_0+ N_k$: the scalar precision in NW distribution. \\
    $\bm{m}_k=\frac{1}{\beta_k}(\beta_0\bm{m}_0+N_k\bar{z}_k)$: the posterior mean of cluster $k$. \\
    $W_k^{-1} = W_0^{-1} + N_k S_k + \frac{\beta_0 N_k}{\beta_0 + N_k}(\bar{\bm{z}}_k- \bm{m}_0)(\bar{\bm{z}}_k- \bm{m}_0)^T$.\\ 
    $\nu_k = \nu_0+N_k$: the $k$th posterior degrees of freedom of NW. \\
    $\bm{\phi}$: variational parameters of the $k$th NW components.\\
   $\eta_t$: variational parameters of a Beta distribution for the\\
   $t$th component in Equation~\ref{eq:variational}.\\
  $\zeta_t$:variational parameters of the NW distribution for $\bm{\phi}_t$. \\
  $\rho_n$: the variational parameters of a categorical distribution \\
  for the cluster membership for each observation.\\
   \bottomrule[\heavyrulewidth] 
   \end{tabular}
\end{table}

\subsection{Benchmark Datasets Description}
\begin{itemize}
\item \textbf{MNIST}: The MNIST dataset consists images of 70000 handwritten digits of $28\times28$ pixel size. In order to compare fairly with previous methods, we did normalization and flattened each image to a $784\times1$ vector.
\item \textbf{STL-10}: The STL-10 dataset consists of color images of $96\times96$ pixel size. There are 10 classes with 1300 examples each. Following previous works, we fed original images to ResNet (\citet{he2016deep}) pretrained on ImageNet (\citet{deng2009imagenet}) and used the last feature map after the $3\times 3$ average pooling layer. So the extracted feature is of size $2048\times1$.
\item \textbf{REUTERS}: The Reuters dataset contains about 810000 English news stories labeled with a category tree. Following previous works, we just used four root categories corporate/industrial, government/social, markets, and economics as labels and discard articles have multiple labels to get 685071 articles. We then randomly sampled a subset of 10000 articles called call REUTERS-10K. As our method are scalable by its online nature, we mainly experimented on REUTERS-10K.
\end{itemize}
\begin{table}[!htp]
\caption{Summary statistics for benchmark datasets.}
\label{tb:DATASET}
\small
\centering
 \begin{tabular}{|| c| c c c ||} 
 \hline
Dataset & \# Samples & Dimension & Classes \\
 \hline\hline
 MNIST & 70000 & 784 & 10 \\ 
 \hline
 REUTERS-10K & 10000 & 2000 & 4 \\
 \hline
 STL-10 & 13000 & 2048 & 10 \\ 
 \hline 
\end{tabular}
\end{table}

\subsection{Evaluation metrics}\label{sec:metric}
\begin{itemize}
    \item \textbf{Normalized Mutual Information (NMI)} is a normalized metric for determining the quality of clustering. It can be used to compare different clusterings with different number of clusters. Its range is between zero and one, which represents no mutual information and perfect correlation. The NMI is defined as follows:
    
    \[
    \textrm{NMI}(l,c) = \frac{2*I(l,c)}{[H(l)+H(c)]},
    \]
    where $l$ is the ground-truth label, $c$ is the cluster assignment by the algorithm, $I$ and $H$ represents mutual information and entropy respectively (the definition for $I$ and $H$ is the same among all the following metrics).

    \item \textbf{Adjusted Rand Index (ARI)} ranges between zero and one. If it is close to zero, it represents random labeling independently of the number of clusters and samples; it equals to one when the clusterings are identical as the true one (up to a permutation). 
    
     Given a set $S$ of $n$ samples, where $C$ is the set of true classes, $C=\{c_{i}|i=1,\dots,n_c \}$ and $K$ is the set of clusters, $K=\{k_{i}|i=1,\dots,n_k \}$. Define $A$ to be the contingency table produced by the clustering algorithm such that every element $a_{ij}$ in $A$ represents the number of samples that are members of class $c_i$ and elements of cluster $k_j$.
     Therefore, the ARI is defined as follows according to \citet{hubert1985comparing}:
     
   \begin{equation*}
   \textrm{ARI} = \frac{{\sum_{ij} \binom{a_{ij}}{2}} - {[\sum_i \binom{a_i}{2} \sum_j \binom{b_j}{2}] / \binom{n}{2}}}{{\frac{1}{2} [\sum_i \binom{a_i}{2} + \sum_j \binom{b_j}{2}]} - {[\sum_i \binom{a_i}{2} \sum_j \binom{b_j}{2}] / \binom{n}{2}}},
\end{equation*}
where $a_{i} = \sum_{j}a_{ij}$ and $b_{j} = \sum_{i}a_{ij}$.
     
    \item  \textbf{Homogeneity Score (HS)} is a homogeneity metric of a cluster labeling given the ground truth. A clustering satisfies homogeneity (with value one) if all of its clusters contain only data points which are members of a single class.
    
    In \citet{rosenberg2007v}, they assume $n$ is number of observations and share the same definition of $C$, $K$ and $A$ as in ARI. They define homogeneity as:

\[
h=\begin{cases}
1 & \textrm{if}\quad H(C,K)=0, \\
1-\frac{H(C|K)}{H(C)} & \textrm{else},
\end{cases}
\]

\[
H(C|K) = -\sum_{k=1}^{|K|}\sum_{c=1}^{|C|}\frac{a_{ck}}{N}\log \frac{a_{ck}}{\sum_{c=1}^{|C|}a_{ck}}
\]
\\
\[
H(C) = -\sum_{c=1}^{|C|}\frac{\sum_{k=1}^{|K|}a_{ck}}{n}\log\frac{\sum_{k=1}^{|K|}a_{ck}}{n} 
\]

Since $H(C|K) \leq H(C)$, the value of $h$ is between zero and one. In the degenerate case where $H(C) = 0$, they define $h$ to be 1.

    \item \textbf{V-measure score (VM)} is a metric to measure the agreenment of two independent clusterings on the same dataset. Its range is between zero and one where one stands for perfect complete clustering as the ground truth. V-measure is the weighted harmonic mean of homogeneity and completeness. \cite{rosenberg2007v} define the completeness measure as follows, which is symmetrical to homogeneity defined as HS previously (definitions of parameters are the same as in HS):
    \[
c=\begin{cases}
1 & \textrm{if}\quad H(K,C)=0, \\
1-\frac{H(K|C)}{H(K)} & \textrm{else},
\end{cases}
\]
where 
\[
H(K|C) = -\sum_{c=1}^{|C|}\sum_{k=1}^{|K|}\frac{a_{ck}}{N}\log \frac{a_{ck}}{\sum_{k=1}^{|K|}a_{ck}}
\]
\\
\[
H(K) = -\sum_{k=1}^{|K|}\frac{\sum_{c=1}^{|C|}a_{ck}}{n}\log\frac{\sum_{c=1}^{|C|}a_{ck}}{n} 
\]
 Similarly, in the degenerate case where $H(K) = 0$, they define $c$ to be 1. \\
 The V-measure is defined as follows:
\[
V_{\beta} = \frac{(1+\beta)*h*c}{(\beta*h)+c},
\]
where $\beta$ is the weighting factor. Note that if $\beta$ is greater than one, completeness is weighted more strongly; if $\beta$ is less than one, homogeneity is weighted more strongly.
\end{itemize}

\subsection{Birth and Merge Strategies}
Our birth and merge strategy is the same as \citet{hughes2013memoized}. 
It is challenging to give birth to new components in a streaming setting since each mini-batch of data may not be sufficient to inform good proposals for new clusters even though the whole data stream may support the new cluster. To resolve this issue, we adopt the same birth strategy as \citet{hughes2013memoized}. 

We first collect a subsample of data for each single cluster $k$. Then we visit each mini-batch in turn and cache samples in this subsample if the probability $\hat{\gamma}_{nk}$ of the $n$th observation to be assigned to cluster $k$  is bigger than a threshold $0.1$. After the collection stage, we fit the DPMM to the collected subsample with $K'=10$ components and run a limited number of variational inference iterations. Then we expand the current model from $K$ clusters to $K+K'$ clusters. We visit each mini-batch of the data stream and perform local and global parameter updates for the expanded model with $K + K'$ clusters. The merge moves have two key steps: (1) select candidate clusters to merge (2) merge two selected clusters if ELBO improves. The candidate clusters are selected based on the ratio of the marginal likelihood of the configuration if the two clusters get merged and the marginal likelihood of the configuration if we keep the two candidates as two separate clusters. We selected the pairs of clusters among those with relatively large marginal likelihood ratio values. 

\subsection{Generated Images from Sequentially Trained AdapVAE}
\begin{figure}[h]
   \centering
	\includegraphics[scale=0.32]{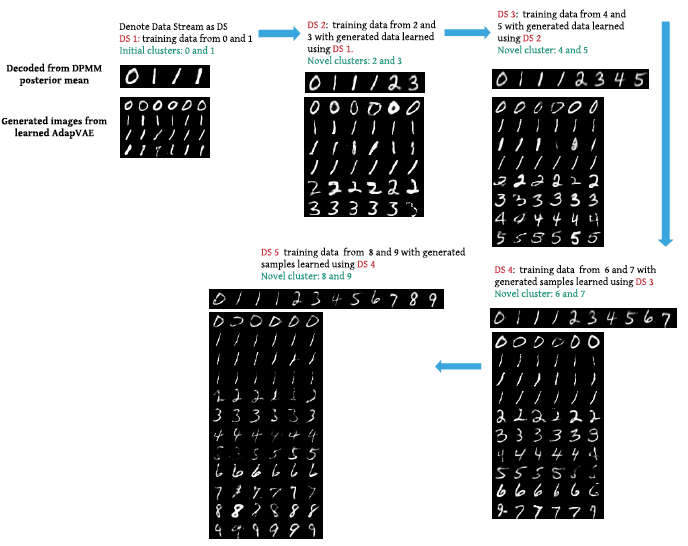}
	\caption{ Decoded image of the posterior DPMM mean parameters and generated images from sequentially learned AdapVAE of Section 6.3.}
    \label{fig:seq}
\end{figure}


\bibliographystyle{apalike}
\bibliography{refs}

\end{document}